\documentclass[conference]{IEEEtran}
\IEEEoverridecommandlockouts
\usepackage{cite}
\usepackage{amsmath,amssymb,amsfonts}
\usepackage{algorithmic}
\usepackage{graphicx}
\usepackage{textcomp}
\usepackage{xcolor}

\def\BibTeX{{\rm B\kern-.05em{\sc i\kern-.025em b}\kern-.08em
    T\kern-.1667em\lower.7ex\hbox{E}\kern-.125emX}}

\usepackage{microtype}
\usepackage{subfigure}
\usepackage{braket}
\usepackage{gensymb}

\begin{document}

\title{Investigating Enactive Learning for Autonomous Intelligent Agents}

\author{
    \IEEEauthorblockN{Rafik Hadfi}
    \IEEEauthorblockA{School of Psychological Sciences,\\ Monash University, Australia\\ rafik.hadfi@monash.edu}
}

\maketitle

\begin{abstract}
The enactive approach to cognition is typically proposed as a viable alternative to traditional cognitive science. Enactive cognition displaces the explanatory focus from the internal representations of the agent to the direct sensorimotor interaction with its environment. In this paper, we investigate enactive learning through means of artificial agent simulations. We compare the performances of the enactive agent to an agent operating on classical reinforcement learning in foraging tasks within maze environments. The characteristics of the agents are analysed in terms of the accessibility of the environmental states, goals, and exploration/exploitation tradeoffs. We confirm that the enactive agent can successfully interact with its environment and learn to avoid unfavourable interactions using intrinsically defined goals. The performance of the enactive agent is shown to be limited by the number of affordable actions.
\end{abstract}


\section{Introduction}
\label{Introd}
The enactive paradigm has originally emerged from embodied cognitive science and particularly from the early work of Maturana and Varela \cite{valera1991embodied,maturana1991autopoiesis}. According to this paradigm, a living organism depends constitutively on its living body and places sensorimotor interactions at the centre of cognition \cite{georgeon2013radical}. Cognition becomes therefore aligned with the organisational principles of living organisms while giving a major role to the phenomenology of experience. The paradigm challenges the separation between the internal constituents of a system and its external conditions, and emphasises the interaction between the two. It also perceives an organism as an autonomous and active entity that is able to adaptively maintain itself in its environment \cite{abramova2017enactive}. For this type of interactions to happen enactivism posits that the agent must be a part of reality \cite{gallagher2013making,barandiaran2017autonomy}.

The paradigm has influenced a large number of embodied cognition theorists \cite{clark1997being,hutto2012radicalizing} and has contributed to the emergence of a variety of research programs such as evolutionary \cite{harvey2005evolutionary} and epigenetic \cite{berthouze2003epigenetic} robotics. In the area of Artificial Intelligence (AI), it is becoming more and more accepted as a viable alternative to the computationalist approaches in building artificial agents that can behave in a flexible and robust manner under dynamic conditions \cite{di2014enactive}. Nevertheless, some concerns have been raised with regard to the sufficiency of the current enactive AI for advancing our understanding of artificial agency and providing accurate models of cognition \cite{froese2009enactive,torranceinter}. The aim of this paper is to provide some initial steps towards the development of such an understanding. We think that enactive cognitive science can provide the conceptual tools that are needed to diagnose more clearly the limitations of current enactive AI, particularly at a time where Reinforcement Learning (RL) is by far the dominant paradigm \cite{russell2016artificial,silver2016mastering,mnih2015human}. The development of an enactive AI would provide fuller models of Enactive Learning (EL) and would challenge the RL methodologies that fail at many real-world problems.

There were few attempts to operationalise enactive cognition in the context of autonomous agents and agent learning. For instance, the authors in \cite{georgeon2015modeling} use the enactive principles to model biological agents. Such agents try to perform rewarding interactions with their environment instead of trying to reach rewarding states as it is the case with RL. In another work, \cite{georgeon2013enactive} formalise the enactive types of interactions between the agent and its environment using an enactive redefinition of Markov Decision Processes. The framework describes a viable architecture that could be used in designing enactive agents but does not evaluate the paradigm in the face of environmental complexity nor it compares it to RL. Departing from the same theoretical framework, we propose to scrutinise enactive learning by building an artificial agent that could act based on enactive principles. We compare such agent to a classical RL agent within complex and volatile maze environments. We then analyse and discuss the different behavioural characteristics of both agents in light of limited access to the environmental states, goals, and exploration/exploitation tradeoffs. We show that enactive the agent can successfully learn to interact with its environment and exploit regularities of sensorimotor interactions. Particularly, it learns to avoid unfavourable actions using intrinsically defined goals.

%The authors in \cite{georgeon2017little} present a pedagogical game that aims at presenting the basics of constructivist learning and developmental AI. 

The paper is structured as following. In section 2, we provide the theoretic foundation of enactive cognition. In section 3, we give the formal agent models as well as the process of enactive learning. In section 4, we provide the methodology and the experimental setting. In section 5, we provide the results and discussion. Finally, we conclude and highlight the future work.

\section{Enactive Cognition}

Enactive cognition is fundamentally compliant with the constructivist school of thought, which perceives learning as creating meaning from experience \cite{ertmer2013behaviorism}.  On the contrary of cognitive psychologists who think of the mind as a reference tool to the real world, constructivists view the mind as filtering input from the world to produce its own unique reality \cite{jonassen1991evaluating}. The key concepts behind the enactive paradigm is that the agent must discover and learn to exploit the regularities in its interactions with the environment \cite{georgeon2013enactive}. Regularities are patterns of sensorimotor interactions that occur consistently and depend on the active interactions between the agent and its environment.

\begin{figure} [ht]
\centering
  \subfigure[Reinforcement Learning] 
  {\includegraphics[height=0.85in]{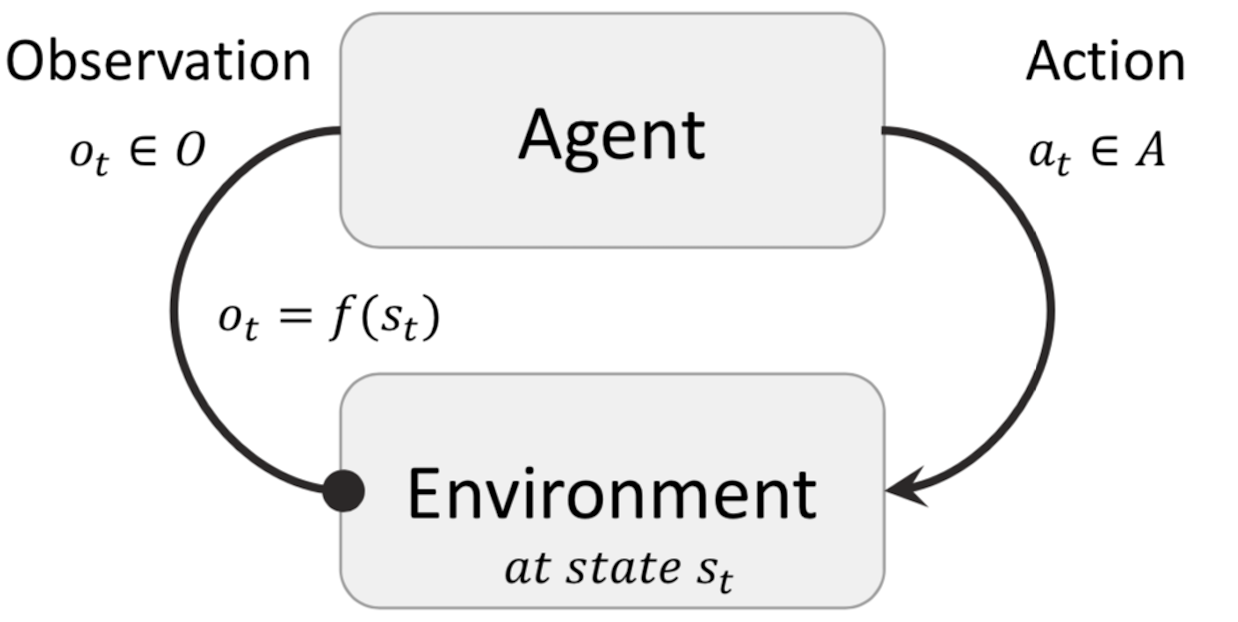}\label{cogparaClass}}
  \subfigure[Enactive Learning]
  {\includegraphics[height=0.85in]{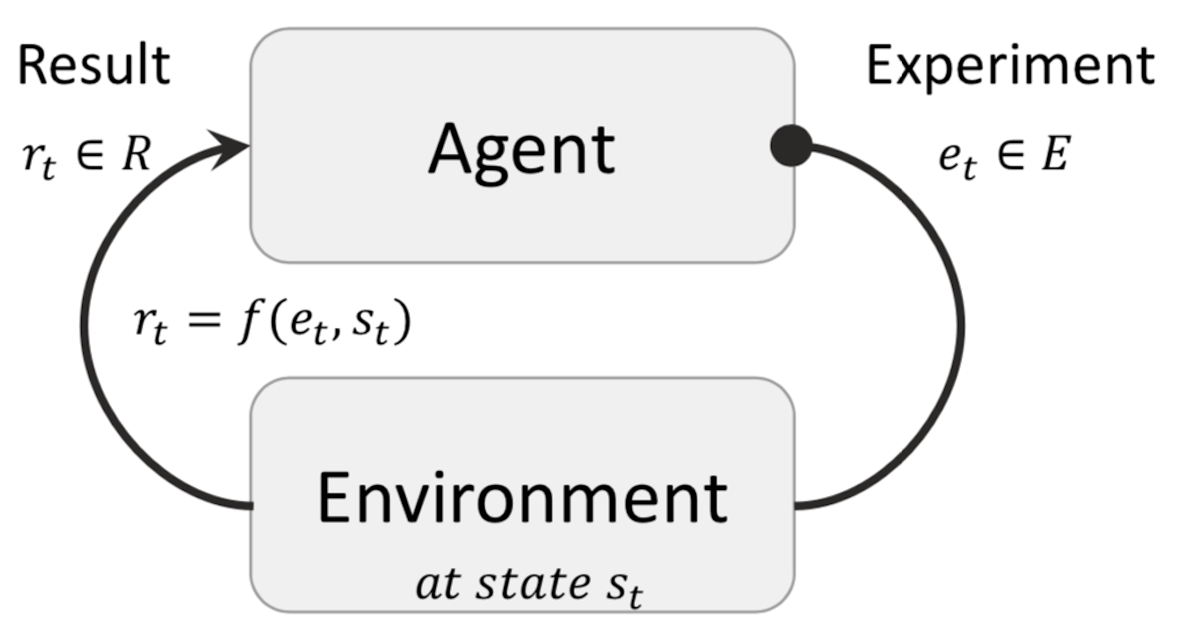}\label{cogparaEnac}}
\caption{Interactions between the agent and the environment in reinforcement and enactive learnings.}
\label{cogpara}
\end{figure}

To better understand how an enactive agent operates, let us contrast it with the type of agent that are often used in RL and rely heavily on internal representations of their environment. Such agent (namely RL agent), passively interprets input data as if it represented the environment. For instance, we could take the situation of a robot exploring a real-world environment. The position of the robot reflects its (spatial) state as seen by an external observer, and its internal state as if the robot is keeping track of its own position. The state of the environment is also available to the robot and would for instance account for the percepts that are available within its visual field. Moving around implies a change in the perceptual field and therefore the state of the environment. Such assumptions do not hold for the enactive agent which is not a passive observer of reality but constructs its perception of the environment through the active experience of interaction \cite{georgeon2017little,georgeon2015constructing}. In this case, the states of the environment are not directly available and the agent is actively involved in shaping its perceptions of those states. We illustrate the distinctions between the two types of agents in figure \ref{cogpara}. We particularly look at how interactions are initiated between the agent and the environment as a succession of decision cycles.

In figure \ref{cogparaClass}, the interaction cycle starts with an observation $o_t$ and ends with an action $a_t$ on the environment. In figure \ref{cogparaEnac}, the cycle starts with the agent performing an experiment $e_t$ and ends by the agent receiving the result $r_t$ of the experiment. 

%Here, the concepts of primitive experiments and results are replaced with the concepts of intended and enacted interactions ($I$). Such substitution departs from the embodiment paradigm to radical interactionism and places interactions at the core of the model \cite{georgeon2013radical}.

Let us now look at the cycles of figure \ref{cogparaEnac}. At the beginning of each decision cycle $t$ of the enactive interactions, the agent decides on an intended sensorimotor interaction to try to enact with reference to the reactive part played by the environment. That is, the agent enacts an interaction $i_t = \braket{e_t, r_t}$ at time $t$, with $i_t$ being an element of the set of primitive interactions $I$. Enacting $i_t$ means experimenting $e_t$ and receiving a result $r_t$. Then, the agent records the two-step sequence $\braket{i_{t-1}, i_t}$ made by the previously enacted interaction $i_{t-1}$ of $i_t$. The sequence of interactions $\braket{i_{t-1}, i_t}$ is called a composite interaction. The interaction $i_{t-1}$ is called $\braket{i_{t-1}, i_t}$'s pre-interaction, noted as $pre(\braket{i_{t-1}, i_t})$, and $i_t$ is called $\braket{i_{t-1}, i_t}$'s post-interaction and is noted as $post(\braket{i_{t-1}, i_t})$. The set of composite interactions known by the agent at time $t$ is defined as $K_t$ and the set $J_t = I \cup K_t$ is the set of all interactions known to the agent at time $t$. When enacted, the primitive interaction $i_t$ activates previously learned composite interactions as it matches their pre-interaction. For example, if $i_t = a$ and if the composite interaction $\braket{a, b}$ has been learned before time $t$, then the composite interaction  $\braket{a, b}$ is activated, which means that it is recalled from memory. Activated composite interactions propose their post-interaction's experiment, in this case: $b$'s experiment. If the sequence  $\braket{a, b}$ corresponds to a regularity of interaction, then it is probable that the sequence  $\braket{a, b}$ can be enacted again. Therefore, the agent can anticipate that performing $b$'s experiment will likely produce $b$'s result. The agent can thus base its choice of the next experiment on this anticipation.

\section {Agent Learning Models}
In the following, we provide the details of the enactive and reinforcement learning models.

\subsection {Reinforcement Learning}
Reinforcement Learning was inspired by behaviourist psychology as it uses feedback (reinforcement) to modify behaviours in the desired direction \cite{ertmer2013behaviorism}. In practice, an agent is built as to take actions in an environment while maximising some cumulative reward. This is usually formalised using a Markov Decision Process (MDP) within a fully described environment. The MDP is generally represented as a tuple $\braket{S, A, R, \mathcal{T}}$ where $S$ and $A$ are the state and action spaces defined on the environment. The function $R: S \times A \times S \rightarrow \mathbb{R}$ is a reward function that determines how much an agent will be rewarded by taking a given action in a given state. The agent has partial control on the outcomes in its model, which is described by the transition probability function (\ref{eq1}),
\begin {equation} \label{eq1}
	\mathcal{T}(s, a, s') = \mathbb{P}(s_{t+1}=s' | s_t = s, a_t = a)
\end {equation}
with $s, s' \in S$ and $a \in A$. The goal of the agent is to find a policy $\pi: S \rightarrow A$ that maximises the discounted summation of rewards. When the utility of each state converges, we get the optimal policy $\pi^{*}$. The optimal policy is found by iterating using the value function $V^{\pi}$ described in (\ref{eq2}),
\begin {equation} \label{eq2}
V^{\pi}(s) = \sum_{s' \in S} \mathcal{T}(s, \pi(s), s') \bigg [R(s, \pi(s), s') + \gamma V^{\pi}(s') \bigg ]
\end {equation}
with $\gamma$ being a discounting factor in $[0, 1]$. The optimal behaviour of the agent is to select an action at each state according to $\pi^{*}$. An optimal behaviour would be a sequence of actions yielding a sequence of occupied states.
 
In the following, we choose to restrict certain features of RL in order to allow for fair comparison with the enactive learning. For $S=\mathbb{R}^2$, we limit the number of states that are available to the agent by defining its scope as distance $\delta \in \mathbb{R}$. The new state space becomes $S_\delta=\{s' \in \mathbb{R}^2\ :\  \lVert s - s' \rVert \leq \delta \}$ with state $s$ being the location of the agent. We also parametrise the exploratory and exploitative behaviours of the agent by a parameter $\alpha \in [0,1]$. $\alpha$ is the probability of using a purely exploratory strategy (random walk) while $1-\alpha$ is the probability of following the optimal policy $\pi^{*}$.
 
\subsection {Enactive Learning}
\subsubsection {Enactive MDP}
\begin{figure}[ht]
        \begin{center}
            \centerline{\includegraphics[height=1.8in]{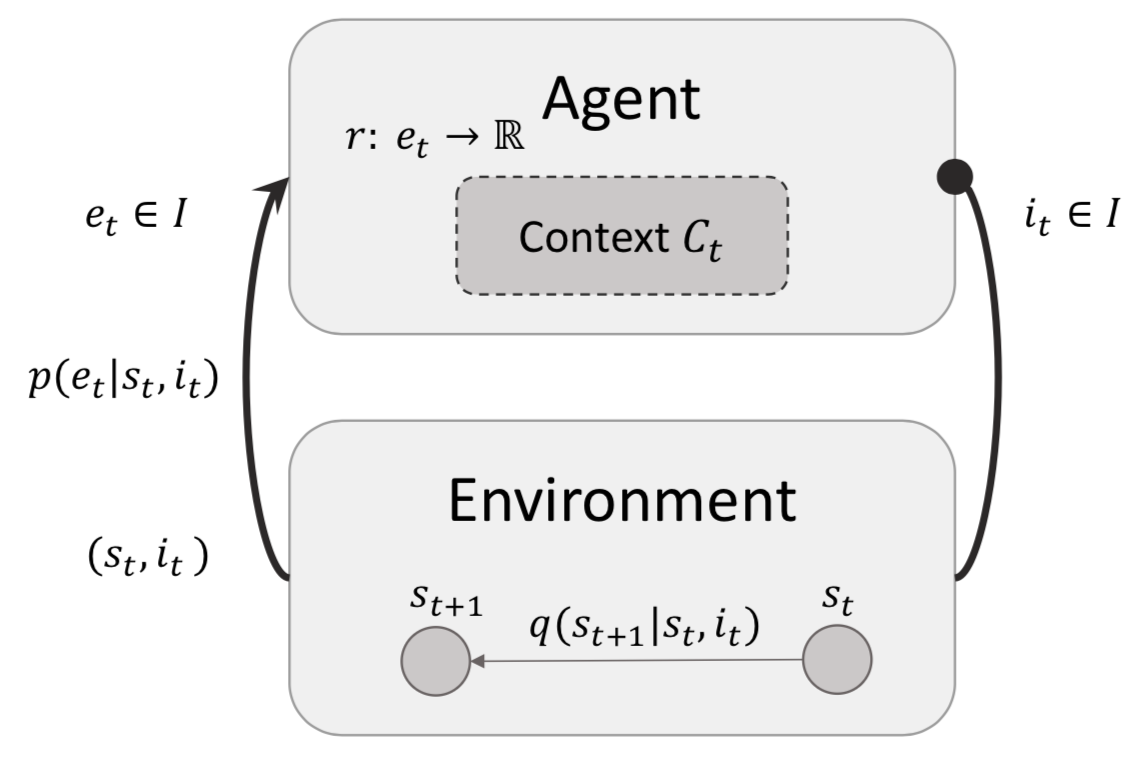}}
            \caption{Enactive Markov Decision Process}
            \label{emdp}
        \end{center}
\end{figure}

The main distinction between RL and the EL resides in the nature of the interactions between the agent and its environment. Such interactions are based on the unique element of actions and results. We can model such interactions using an Enactive Markov Decision Process (EMDP) as shown in figure \ref{emdp}. An EMDP \cite{georgeon2015modeling} is defined as a tuple $\braket{S, I, q, p}$ with $S$ being the set of environment states; $I$ the set of primitive interactions offered by the coupling between the agent and the environment; $q$ a probability distribution such that $q(s_{t+1} | s_t, i_t)$ gives the probability that the environment transitions to state $s_{t+1} \in S$ when the agent chooses interaction $i_t \in I$ in state $s_t \in S$; and $p$ is a probability distribution such that $p(e_t | s_t , i_t)$ gives the probability that the agent receives input $e_t \in I$ after choosing $i_t$ in state $s_t$.

In the EMDP, $i_t$ is called the intended interaction as it represents the sensorimotor scheme that the agent intends to enact at the beginning of step $t$; and constitutes the agent's output that is sent to the environment. We call $e_t$ the enacted interaction because it represents the sensorimotor scheme that the agent records as actually enacted at the end of step $t$; $e_t$ constitutes the agent's input received from the environment. If the enacted interaction equals the intended interaction ($e_t = i_t$) then the attempted enaction of $i_t$ is considered a \textit{success}, otherwise, it is considered a \textit{failure}.

\subsubsection {The Learning Process}

The mechanism underlying the EMDP could be implemented as a sequential learning process that relies on the interactions between the agent and its environment \cite{georgeon2015modeling}. The enactive learning process operates at every decision step $t$ according to the following 7 phases.

\begin {enumerate}
\item \textbf{Preparation}: The agent is initially presented with a set of interactions $C_t \subset J_t$ referred to as the context, with $C_0 = \emptyset$ and $J_0 = I$.

\item \textbf{Activation}: The agent takes the previously learned composite interactions whose pre-interaction belongs to the current context and activates them, forming the set $A_t$ of activated interactions defined as $A_t = \{ a \in K_t | pre(a) \in C_t \}$.

\item \textbf{Proposition}: The activated interactions in $A_t$ propose their post-interaction for enaction, forming the set $P_t$ of proposed interactions defined as $P_t = \{ p \in J_t | \exists a \in A_t, p = post(a)\}$.

\item \textbf{Selection}: The intended interaction $i_t$ is selected from the proposed interactions in $P_t$ based on the proclivity of the interactions. The proclivity of an interaction $i_t$ is defined as $proclivity(i_t) = r(i_t) \times \mathbb{P}(i_t)$ and reflects the regularity of the interaction based on its probability of occurrence and the motivations of the agent.

The selection of the intended interaction is also subject to parameters $\alpha \in [0,1]$ and $d \in \mathbb{R}$. $\alpha$ is the probability that $i_t$ is selected randomly for exploratory purposes, and $1-\alpha$ is the probability that $i_t$ is selected based on proclivity. The parameter $d$ encodes the limited foresight of the agent and specifies how deep it can go in the hierarchy of interactions. In other words, $d$ is the length of the allowable sequences of interactions.

\item \textbf{Enaction}: The agent tries to enact the intended interaction $i_t$, which could (or not) result in an enacted interaction $e_t$.

\item \textbf{Learning}: New composite interactions are constructed or reinforced with their pre-interaction belonging to the context $C_t$ and their post interaction being $e_t$, forming the set of learned or reinforced interactions $L_t$ to be included in $K_{t+1}$. The set $L_t$ is defined as $L_t = \{\braket{pre(i), e_t}\}$.

\item \textbf{Construction}: A new context $C_{t+1}$ is constructed to include the stabilized interactions in $e_t$ and $post(e_t)$:  $C_{t+1} \leftarrow L \eta_t \cup \{e_t\} \cup \{ post(e_t)\}$.
\end {enumerate}

In the following section, we propose to test the enactive and reinforced mechanisms within artificial agents.

\section{Methodology and experimental scenario}

\begin{figure}[ht]
\begin{center}
\centerline{\includegraphics[height=1.2in]{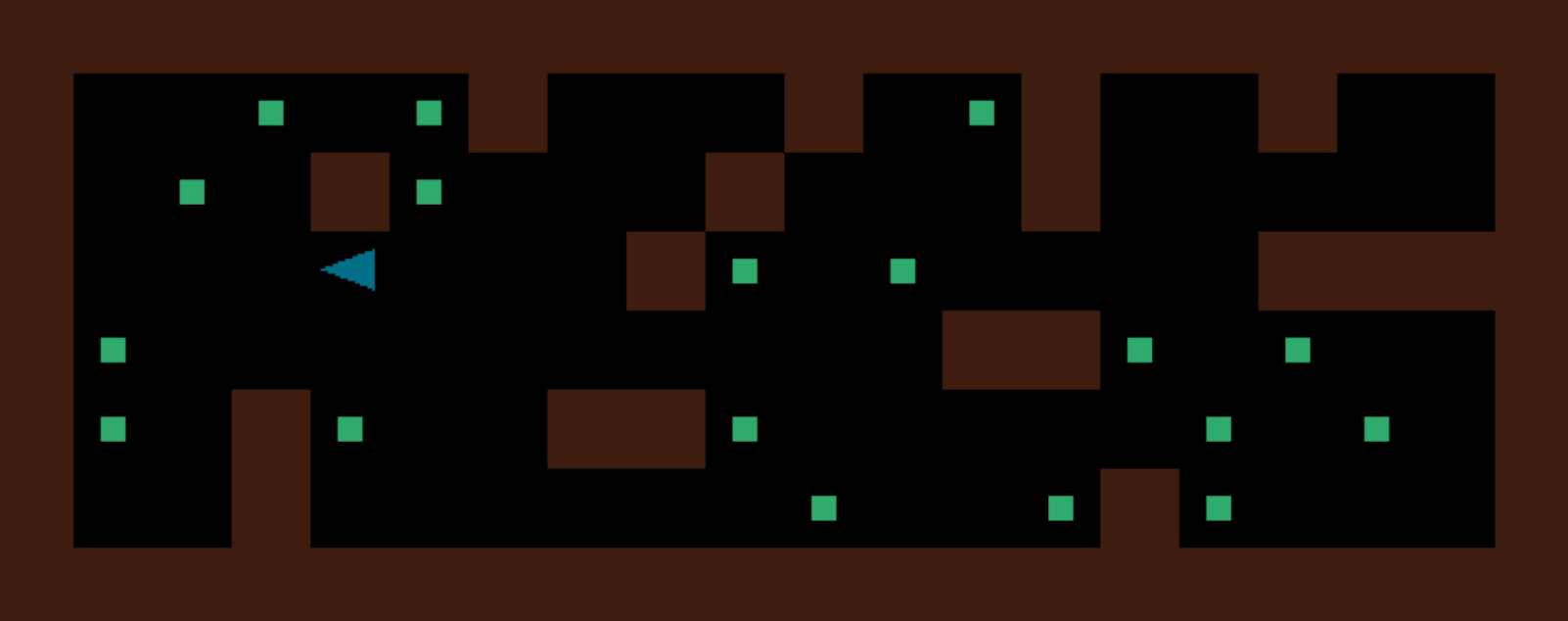}}
\caption{Maze with one agent and 18 food units}
\label{mazAagent}
\end{center}
\end{figure}

To evaluate the two paradigms we set up an experimental scenario in which two agents are expected to perform a foraging task in a 2D maze environment as illustrated in figure \ref{mazAagent}. Each agent is tested on its own and both agents start from the same position.

\subsection {The environment}
 The environment is a $8 \times 10$ maze and is defined in terms of its structure and behaviour. The structure of the maze reflects the difficulty of the problem based on the existence of obstacles and how they limit the access to the food units. The obstacles and their distribution are important when we evaluate the exploratory behaviours of the agents. The dynamic aspects of the environment reside in uniformly adding 20 food units every 200 ticks for an overall period of 1000 ticks per trial.

\subsection {The agents}
An agent is an entity that moves in a 2D space according to a predefined list of actions: moving forward by one step, turning right by $90^{\circ}$, turning left by $90^{\circ}$. The enactive agents possess few additional actions that correspond to failures of interactions. For instance, the action ``step'' has another action ``step fail'' that corresponds to a failed ``step'' action and happens when the agent is attempting to move forward despite the existence of an obstacle. In the following simulations, we consider two types of agents.

\begin{enumerate}
\item An enactive agent that uses EL and interacts with the environment according to an EMDP.

\item A RL agent that interacts with the observable environment according to an MDP. The agent uses a Q-Learning mechanism by updating the reward states based on any dynamically added food unit.

\end{enumerate}
The rewards of the RL agent are defined as in (\ref{RLfunc}).
\begin{equation} r(s_t) =  \left\{
  \begin{array}{ll}
      +5 & \textit{if\ \ $s_t$ = food cell} \\
      0.04 & \textit{if\ \ $s_t$ = empty cell} \\
      -9 & \textit{if\ \ $s_t$ = obstacle cell} \\
\end{array} 
\right.
\label{RLfunc}
\end{equation}

Since the enactive agent does not possess an extrinsically defined utility function that maps states to rewards \cite{sutton1999between,barto2004intrinsically}, we need to define an intrinsic valuation of its actions. Intrinsic motivational values are the main driver of the agent's spontaneous exploratory behaviour \cite{georgeon2013enactive,oudeyer2009intrinsic,oudeyer2008can}. In particular, we want the rewarding mechanism to account for the adaptive behavioural strategies of the agent in the face of constantly changing environments. We define the agent's intrinsic motivations using a reward function $r: I \rightarrow \mathbb{R}$ that maps primitive interactions to motivational states (\ref{rfunc}).
\begin{equation} r(i_t) =  \left\{
  \begin{array}{ll}
      +10 & \textit{if\ \ $i_t$ = step} \\
      -1 & \textit{if\ \ $i_t$ = step fails} \\
      -0.3 & \textit{if\ \ $i_t$ = turn right} \\
      -0.3 & \textit{if\ \ $i_t$ = turn left}
\end{array} 
\right.
\label{rfunc}
\end{equation}

% (Rever) the choice of feedback values is not justified and seems arbitrary. The underlying intuitions should be clarified.
The feedback values (\ref{rfunc}) should in principle be chosen as to encode the expected behaviour that would lead to the desired goals. In RL, the goal is extrinsically specified as scalar reward values. In EL, goal-oriented behaviours are specified as intrinsic motivational values \cite{georgeon2015modeling}. For instance, the ÒstepÓ action has a fixed positive valence that does not change over time, which means that the agent has the same level of motivation to walk forward. Similarly, turning right and left have low negative valences, which means that the agent would often avoid turning.

\section {Experimental Results}

\subsection {The Enactive Learning}
Quantifying the enactive learning requires at first an evaluation of the agentÕs ability to avoid negative-valence enactions. The valence $V_t$ is the value that the agent assigns to an interaction $i_t$, based on its intrinsic motivational value and its probability of occurrence. The enactive learning translates to learning to choose interactions that have positive valence and avoid the interactions that have negative valence.

\begin{figure}[ht]
    \begin{center}
    \centerline{\includegraphics[height=2.8in]{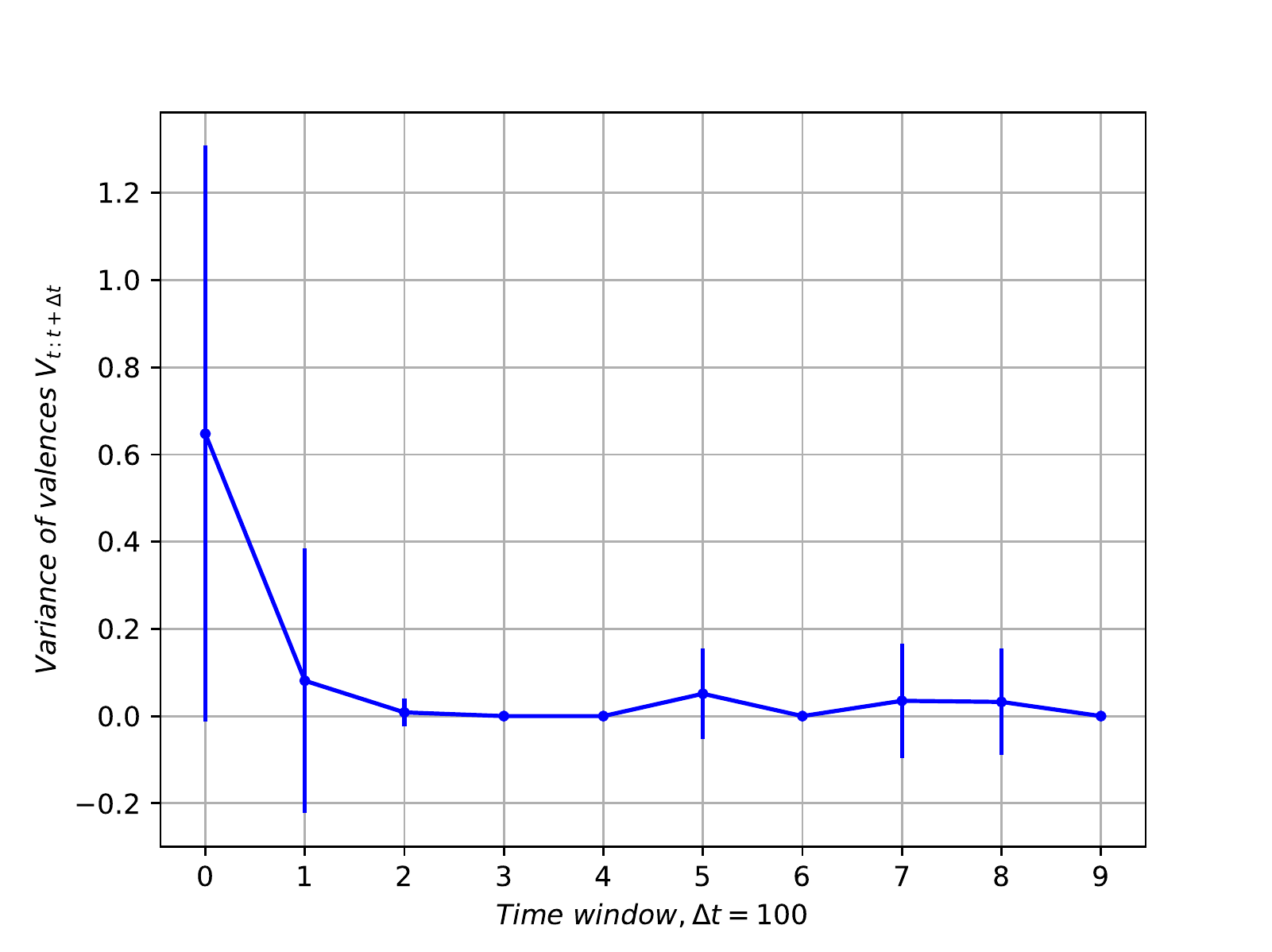}}
    \caption{Reduction in negative valence}
    \label{negValRed}
    \end{center}
\end{figure}

Figure \ref{negValRed} shows the reduction in negative valence counts for 15 runs of the EL agent with $\delta=10$ and $\alpha=0$. We started by dividing the 1000-tick period into 10 time windows of length $\Delta t=100$ each. Then, we calculated the means of all the standard deviations across the 15 trials of each time window.
The reduction of variance reflects a stabilisation in the negative counts, and that actions with negative valence are becoming less and less frequent as time goes on. This also means that the agent has mastered the enactions and their consequences, and in our case, learnt to avoid bumping into the wall while maintaining the combinations of actions (turning left/right) that do have negative valence but are required to explore the environment. Such balance reflects the trade off between exploitation and exploration in the sense that the agent learns to maximise valences.

%Ultimately, the cumulative count of bumps becomes constant as it is shown in figure \ref{bumpCount}. This translates also to a stabilised exploration policy and a reduction in the entropy of limiting space-states distribution as shown in figure \ref{expH}. In other words, the agent starts focusing on the partitions of the maze that yield an optimal behaviour.
%\im{figure of entropy?}

\subsection {Comparing the learning paradigms}

In the following, we compare the foraging performances of the agents for exploratory behaviours $\alpha \in \{0, 0.5\}$ across foresights $d \in [2,20]$ and $\delta \in [2^1, 2^{11}]$.

Since the notion of space is not explicitly defined for EL, we need to interpret how $d$ and $\delta$ map to each other. For RL, an increment in $\delta$ corresponds to an increment in space coverage, for instance, moving from position $x$ to position $y$ happens with $\lVert x - y \rVert \le \delta$. Limiting the choice of interactions with motivational values (\ref{rfunc}), makes the EL agent more likely to pick the ``step'' interaction instead of the others. The interaction ``step'' corresponds in reality to a change in space. Therefore, an increment in $d$ corresponds to an increment in the space coverage. For instance, let us take the composite interaction $\braket{i_t,\braket{i_{t+1}, i_{t+2}}}$ composed of successful ``step'' actions: $e_t=e_{t+1}=e_{t+2}=``step"$ and $r_t=r_{t+1}=r_{t+2}=``Success"$. This sequence causes the agent to change position from $s_t$ to $s_{t+3}$ with $\lVert s_t - s_{t+3} \rVert = 3$. This type of mapping illustrates how internal motivational models map to behaviours. Programming the enactive agent corresponds to finding the right motivational model that would reproduce the desired behaviours.

\begin{figure}[h]
\centering
  \subfigure[Reinforcement Agent] 
  	{\includegraphics[height=2.8in]{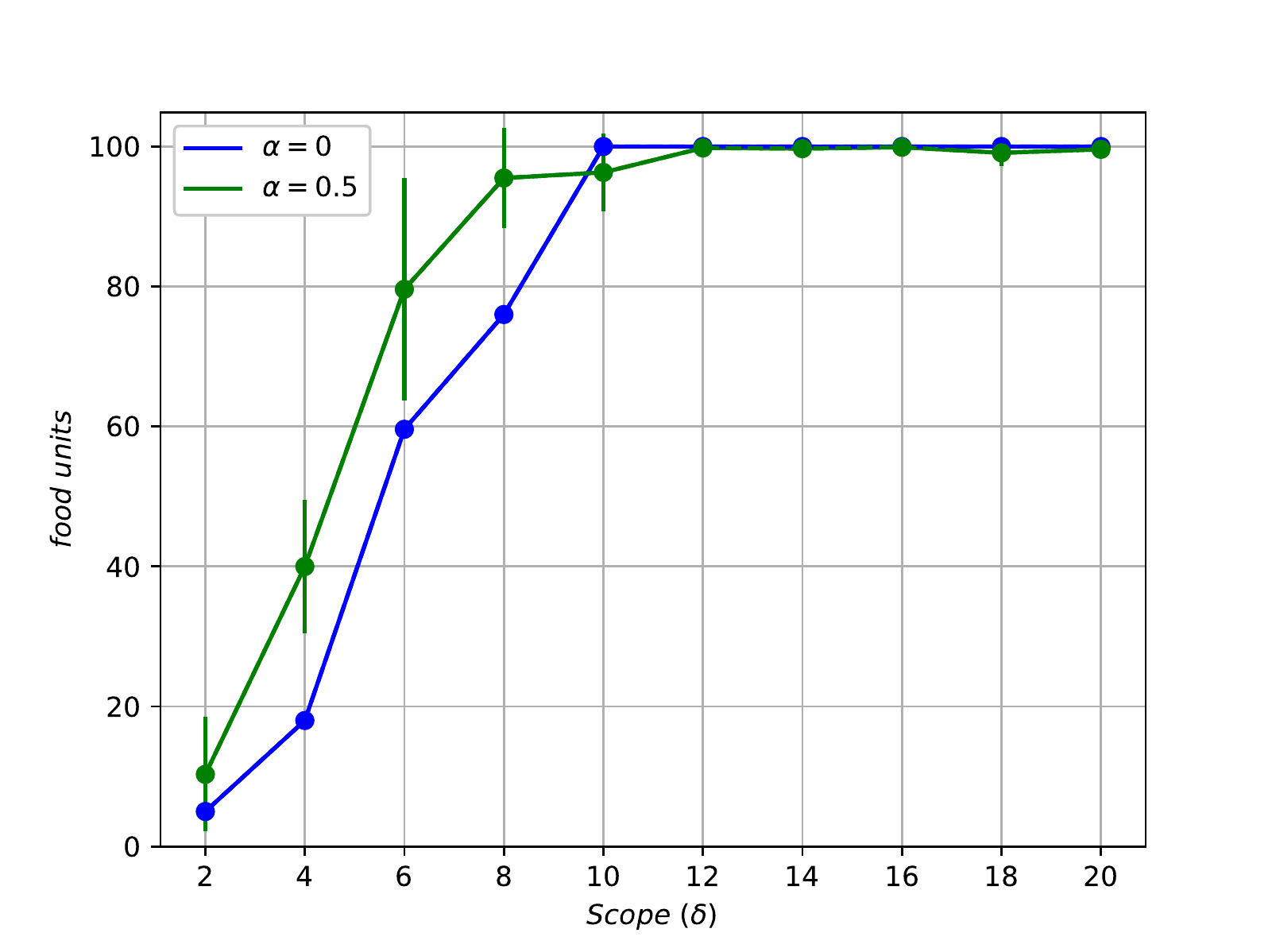}\label{RLfig}}
    \subfigure[Enactive Agent]
    {\includegraphics[height=2.8in]{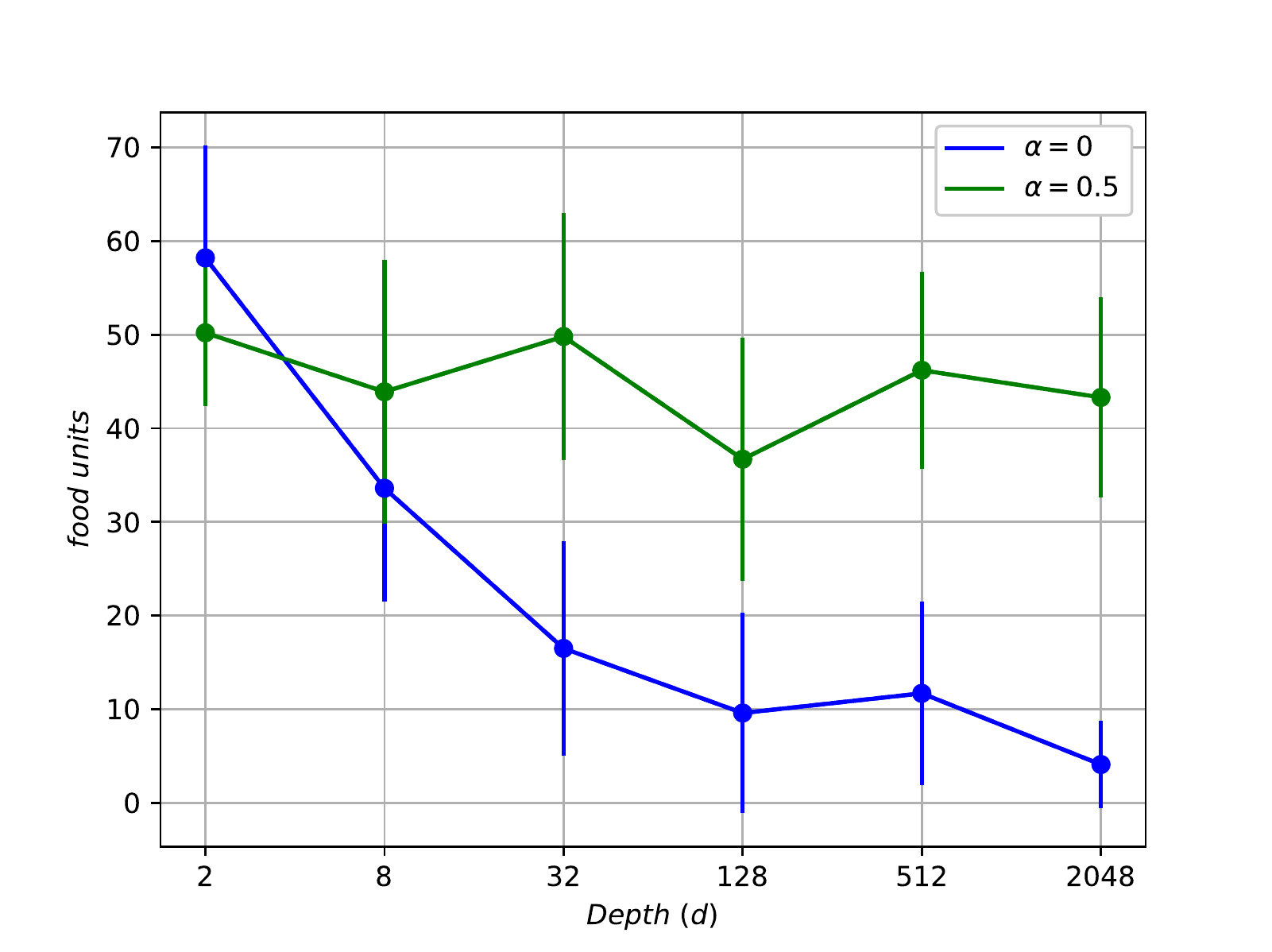}\label{ELfig}}
\caption{Foraging for different exploratory strategies}
\label{PerfomanceFig}
\end{figure}

The performances of the agents are shown in figure \ref{PerfomanceFig}. In figure \ref{RLfig}, the RL agent exhibits a systematic, deterministic behaviour for $\alpha=0$ since it is solely governed by its policy. Adding and explorative behaviour ($\alpha=0.5$) improves the gain when the foresight is limited (scope [2-10]) but goes up for $\delta=10$, which means that the scope covers most of the $8 \times 20$ maze. When the scope covers all the space ($\delta \simeq 20$), the agent's gain is mainly driven by exploitation.

Despite the poor overall performance shown in figure \ref{ELfig}, the EL agent did actually start well for $\delta=2$ with a performance within [50,60] compared to that of the RL agent (20), but then its performance went down with the increase of $d$. The drop of performance of $\alpha=0$ for high $d$ values is mainly due to the usage of long sequences of sensorimotor interactions, which take long to enact, evaluate and learn. Complex enactions impair the agentÕs ability to exploit the space and traps him in suboptimal areas. This is true in our case even with 4 primitive interactions and should grow intractably if we add more interactions. A possible solution to reduce this complexity is to interrupt the activation cycle of long sequences of interactions by randomly picking one primitive interaction instead. The use of such an exploratory behaviour is visible for $\alpha=0.5$ and yields more gain.

Being trapped in complex interactions does not necessarily translate to a goal-oriented behaviour for an external observer but accounts for the spontaneity and the self-motivation that drives the EL agent \cite{gay2017autonomous}. The goal-driven behaviour becomes an emergent property of the constructs (\ref{rfunc}). The challenge is therefore to find the appropriate mapping from rewards (\ref{RLfunc}) to motivational values (\ref{rfunc}) while balancing the exploration/exploitation tradeoff with an optimal $\alpha$. 

Enactive learning could fail at problems that cannot be defined in terms of states either for the lack of any formal description of the state space or when the state space is too large to be encoded in a reward function. While reinforcement learning is more performant when the state space is well defined, it still lacks the ability to maintain its scalability for large spaces. Enactive learning on the other hand has the ability to construct its own map of the state-space and exploit it based on its intrinsic model of behaviour. We summarise the main difference between the two learning paradigms in table \ref{tab1}.

\begin{table*}[h]
\caption{Comparing Enactive and Reinforcement Learning}
\begin{center}
        \begin{tabular}{|c|c|c|}
        \hline
         \textbf{\textit{Criteria}} & \textbf{\textit{Enactive Learning}}& \textbf{\textit{Reinforcement Learning}}\\
        \hline
        \hline
        Theoretical framework & Constructivism & Behaviourism  \\
        \hline
        Formal model & Enactive MDP (EMDP) & MDP \\
        \hline
        Mechanism & Sequential Learning & Value Iteration\\
        \hline
        Reward & Intrinsic, not a function of the environment (self-motivation)& Extrinsic, function of environment\\
        \hline
        Scope & Depth in hierarchies of interactions & Distance on the state space\\
        \hline
        Goal & Emergent, mainly driven by self-motivation & Predefined, using a reward function \\
        \hline
        Representation & Gradient of actions & Hierarchies of interactions\\
        \hline
        Perception & Percepts are internally constructed & Percepts are function of the environment \\
        \hline
        Input/Output & Result/Experiment  & Observation/Action \\
        \hline
        Action cycle & Starts from the agent & Starts from the environment\\
        \hline
\end{tabular}
\label{tab1}
\end{center}
\end{table*}

\section{Conclusion and future work}

Enactive Learning is an interesting alternative to RL given its capability to operate without prior knowledge of the states of the environment. We developed an artificial agent that learns based on the enactive principles, and compared its behaviours and performances to an agent that runs on RL. The agents are tested in foraging tasks within complex and volatile environments. We show that the enactive agent can successfully interact with its environment, and learn to avoid unfavourable interactions using intrinsically defined goals. The enactive agent is shown to be limited by the number of affordable actions that it could enact at a time. Limiting the size of its memory of interactions or relying on exploratory strategies increases its performance.

As future directions, we would like to change the way we defined the intrinsic motivational values as scalar values, and instead use expectations of rewards as it is done in the predictive coding framework of \cite{friston2015discussion}. Moreover, we think of investigating the structures of hierarchies of interactions and whether they could be used to build spatial representations of the environment and of the agent itself.

\newpage

\bibliography{ref}
\bibliographystyle{ieeetran}

\end{document}